\documentclass[conference]{IEEEtran}
\IEEEoverridecommandlockouts

\usepackage{cite}
\usepackage{amsmath,amssymb,amsfonts}
\usepackage{algorithmic}
\usepackage{graphicx}
\usepackage{textcomp}
\usepackage{xcolor}
\usepackage{url}
\usepackage{subfig}
\newtheorem{theorem}{Theorem}[section]

\newtheorem{lemma}[theorem]{Lemma}

\newcommand{\algo}{SPEED}
\def\BibTeX{{\rm B\kern-.05em{\sc i\kern-.025em b}\kern-.08em
    T\kern-.1667em\lower.7ex\hbox{E}\kern-.125emX}}
\begin{document}

\title{{Know What You Don't Know: Selective Prediction for Early Exit DNNs}
\thanks{Divya Jyoti Bajpai is supported by the Prime Minister’s Research Fellowship (PMRF), Govt. of India.  Manjesh K. Hanawal thanks funding support from SERB, Govt. of India, through the Core Research Grant (CRG/2022/008807) and MATRICS grant (MTR/2021/000645), and DST-Inria Targeted Programme.  We gratefully acknowledge the funding support from the Amazon IIT-Bombay AI-ML Initiative (AIAIMLI). We also thank Harsh Jitendrakumar Bundeliya for his assistance with the initial code setup and experiments.
}
}

\author{\IEEEauthorblockN{Divya Jyoti Bajpai and Manjesh Kumar Hanawal}
\IEEEauthorblockA{\textit{Department of IEOR, IIT Bombay, Powai, Mumbai} \\
\{divyajyoti.bajpai, mhanawal\}@iitb.ac.in}
}

\maketitle

\begin{abstract}
Inference latency and trustworthiness of Deep Neural Networks (DNNs) are the bottlenecks in deploying them in critical applications like sensitive tasks.  
Early Exit (EE) DNNs overcome the latency issues by allowing samples to exit from intermediary layers if they attain `high' confidence scores on the predicted class. However, the DNNs are known to exhibit overconfidence, which can lead to many samples exiting early and render EE strategies untrustworthy. We use Selective Prediction (SP) to overcome this issue by checking the `hardness' of the samples rather than just relying on the confidence score alone.  We propose \algo{}, a novel approach that uses Deferral Classifiers (DCs) at each layer to check the hardness of samples before performing EEs. Specifically, the DCs identify if a sample is hard to predict at an intermediary layer, leading to hallucination, and defer it to an expert. Early detection of hard samples for inference prevents the wastage of computational resources and improves trust by deferring the hard samples to the expert. We demonstrate that EE aided with SP improves both accuracy and latency. Our method minimizes the risk of wrong prediction by $50\%$ with a speedup of $2.05\times$ as compared to the final layer. The anonymized source code is available at \url{https://github.com/Div290/SPEED} 
\end{abstract}


\section{Introduction}

The demand for Artificial Intelligence (AI) systems to automate decision-making is growing. However, they face two major issues in deployment: 1) their computational resource requirements and the associated inference latency. 2) In socially sensitive or mission-critical machine learning applications, their trustworthiness is a concern \cite{kaur2022trustworthy}. For instance, the ability to know what you do not know and not get overconfident is essential. However, the DNNs are prone to false overconfidence (fake confidence) \cite{gawlikowski2023survey}, making them vulnerable to wrong decisions and leading to a trust deficit.

\begin{figure}
    \centering
    \includegraphics[scale=0.5]{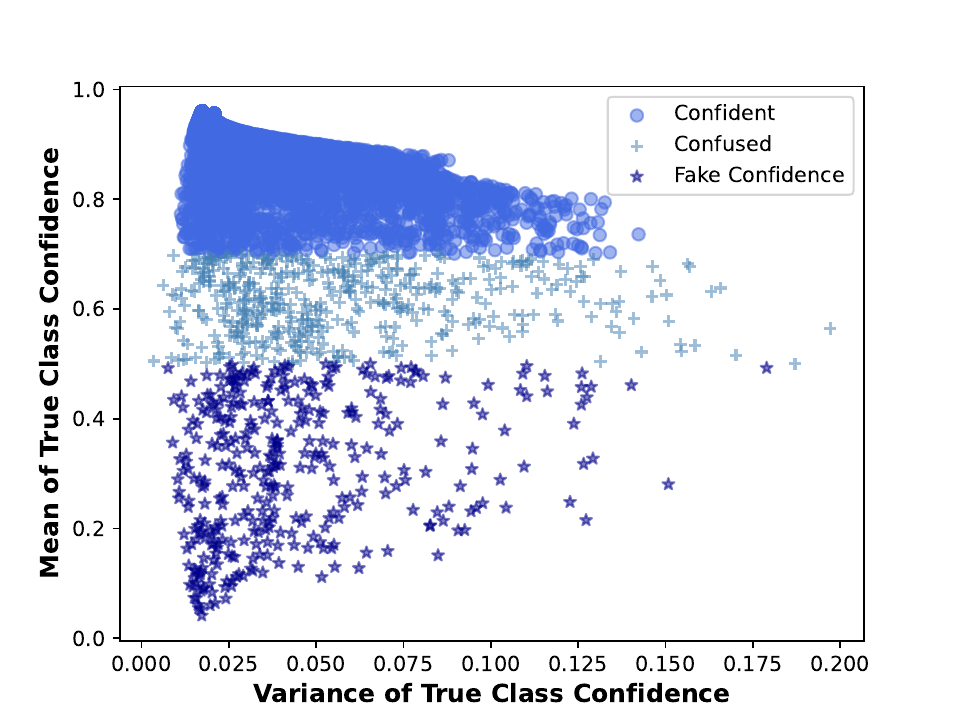}
    \caption{The figure shows a plot of average confidence values across the layers over the true class on the SST-2 dataset.}
    \label{fig:average_confidence}
    \vspace{-0.5cm}
\end{figure}
To address the latency issue, various adaptive inference methods are developed, including Early Exit DNNs (EEDNNs) \cite{kaya2019shallow, zhou2020bert}. EEDNNs use exit classifiers attached to the intermediary layers to perform adaptive inference, allowing samples attaining good confidence scores on the predicted class to exit at shallow layers without requiring them to pass through all layers, thus reducing the average inference latency. They have been widely adopted in resource-constrained scenarios because of their significant inference acceleration with minimal loss in accuracy \cite{10.1145/3639856.3639873, 10622954, pacheco2021early, bajpai-hanawal-2024-ceebert, zhou2020bert}.

However, EEDNNs inherit certain characteristics of standard DNNs, including susceptibility to overconfidence. In fact, they are even more prone to this issue due to the presence of multiple intermediate classifiers that make predictions without utilizing the full depth of the backbone network. Classifiers operating at earlier layers are particularly vulnerable, as they rely on less-refined feature representations, leading to overconfident yet unreliable predictions. This trade-off between inference speed and reliability highlights the necessity of incorporating Selective Prediction (SP) strategies to enhance the trustworthiness of EEDNNs.

Selective Prediction (SP) \cite{chow1970optimum} is a method used in classical DNN to improve the trustworthiness of the model. It enables the model to abstain from making predictions at the final layer when it is uncertain about the true class and instead defer the sample to an expert for further evaluation. This improves the trust in the model. However, existing SP methods improve trust by abstaining from the samples where the model is underconfident but cannot solve the overconfidence issue. Hence, they cannot be directly used in EEDNNs as they typically fail to detect those samples when the model gains fake confidence.

To overcome this, we propose to extend the SP to detect if an exit layer can get overconfident and use an expert to assign a label. The model at every layer decides whether to defer a sample to an external expert (a larger model or a human expert) or send it to the exit classifier. This approach mitigates the risk of incorrect high-confidence predictions, thereby improving the overall trustworthiness of the model. As the EEDNNs have classifiers attached to every layer, the main challenge is to detect samples with fake confidence at each layer, whereas the traditional methods focus on detecting samples with low confidence at the final layer.

Adaptation of traditional SP methods \cite{pugnana2024deep} for EEDNNs is challenging due to two major issues: 1) Existing methods use the confidence score output by the model to decide whether to abstain or not on each sample, which is not a reliable metric to abstain. This is illustrated in Fig~\ref{fig:average_confidence} (details in Sec. \ref{sec: motivation}) for the SST-2 dataset with a task of sentiment classification with \textit{positive} and \textit{negative} classes. We plot the average confidence of the samples in the true class across the layers using the trained EEDNN backbone. It shows that when the true class confidence is small, the samples are `hallucinated' and gain fake confidence at the intermediary layers on the wrong class, making the model predict those samples wrongly with high confidence.  If applied directly, SP methods can get fooled by the model's fake confidence in intermediary layers. 

2) Existing methods determine whether to abstain only at the last layer, i.e., after passing through all the layers, which undermines the speed advantages of EEDNNs. Moreover, the overconfidence issue in DNNs usually manifests in the early layers of the network and grows in deeper layers. For instance, the samples in the lower part of Figure \ref{fig:average_confidence} (labeled as fake confidence) once attained high fake confidence at the shallow layer, rarely revert to true confidence (see Appendix \ref{sec: fake_conf}). Early detection of such samples is useful and can help improve efficiency by preventing unnecessary computation.


\begin{figure*}
    \centering
    \includegraphics[scale = 0.45]{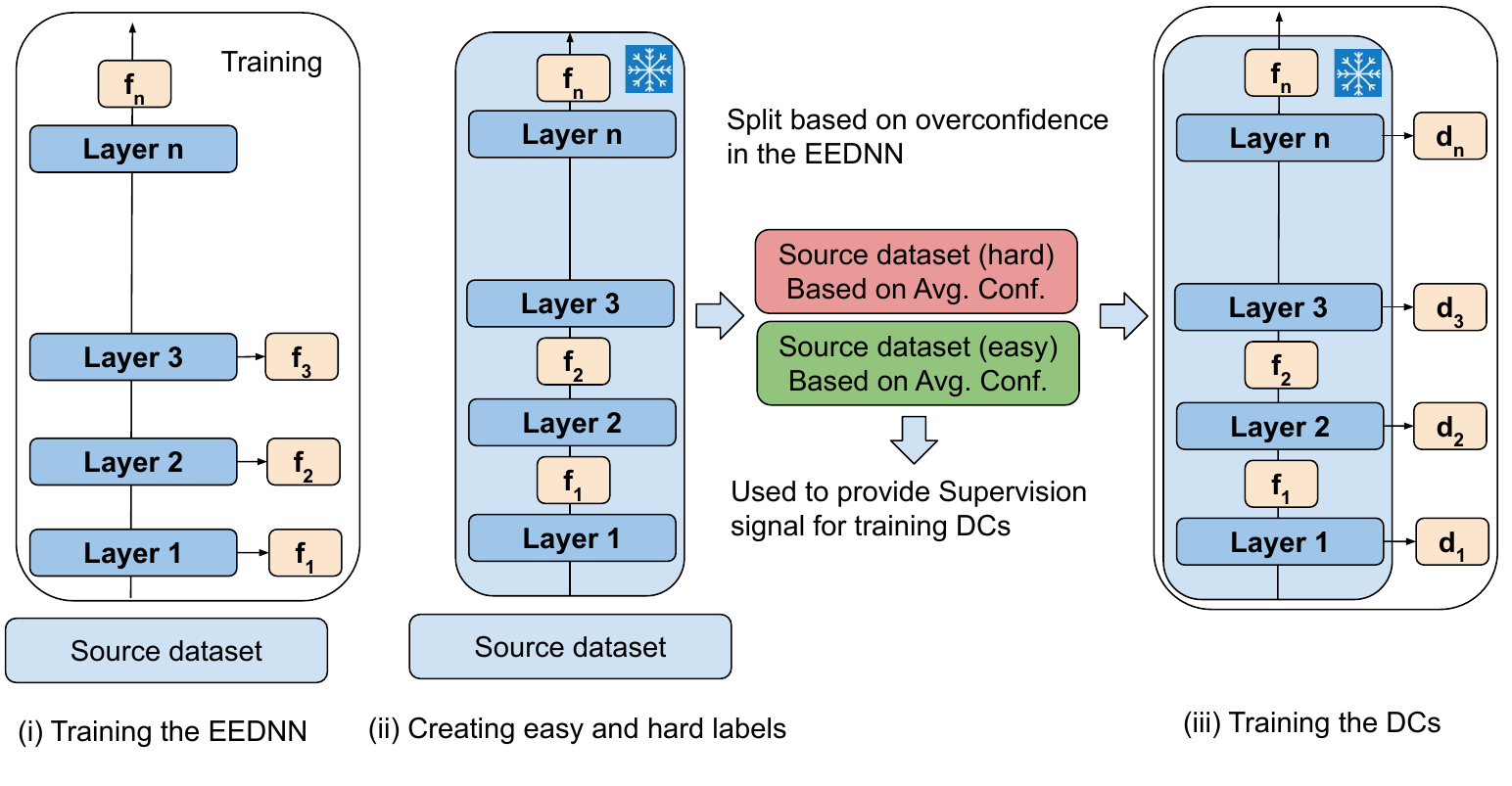}
    \caption{Flowchart of SPEED: i) The EEDNN backbone is trained with attached exit classifiers. ii) Split the dataset into easy and hard using the average confidence obtained from all the exit classifiers. iii) The splits are then used to train the DCs using labels as easy or hard.}
    \label{fig:flowchart}
\end{figure*}



To address these issues, we propose a new framework named \underline{S}elective \underline{P}rediction for \underline{E}arly \underline{E}xit \underline{D}NNs (\algo{}). This framework introduces deferral classifiers (DCs) at each layer of an EEDNN to detect if a sample is hard and the model can get overconfident, thus helping EEDNNs maintain the speed advantage while avoiding making inferences on samples that it does not understand. Our method also identifies samples that attain low confidence at multiple layers and sends them to the expert. Identifying the sample on which the layers can get either overconfident or underconfident early helps in saving computation and faster inference.  To train the DCs, we use samples that are labeled as `easy' or `hard' based on the average confidence they attain across the layer of EEDNN (detain in Subsection \ref{sec:training_DCs}). This makes DCs aware of whether the confidence gained is fake or true. 
During inference, at each layer, the attached DC classifier detects if a sample is easy or hard. If it is hard, it exits the sample from DNN and sends it to an expert. Otherwise it is sent to the Exit Classifier (EC) attached to the same layer. If the EC completes the inference, the sample exits with a label, else it is sent to the next layer where the process repeats. Fig~\ref{fig:flowchart} shows the flowchart of \algo{}.




Our method is robust to domain shifts in text and image classification, as these shifts do not alter sentence semantics or core image characteristics. For example, in sentiment analysis, an IMDB review: “The movie would have been good if there was a better climax!” (negative sentiment) may be misclassified as positive due to the word “good.” Similarly, a hotel review: “The place would have been good if the food quality was better,” can lead to false confidence. Our framework, once trained on a source domain, effectively mitigates such errors in the target domain, reducing the need for retraining.

We also provide a theoretical analysis to develop a condition on the error rate of the DCs such that the overall model performance has a risk lower than a given value. Our findings suggest that the better the DC will be, the lower the risk of wrong prediction by the EEDNN, which motivates us to find a method to train better DCs. We provide results on multiple NLP tasks such as sentiment classification, entailment classification, natural language inference (NLI) tasks and image classification tasks to demonstrate that our method can reduce the risk by $50\%$ while getting an inference speedup of $2.05\times$. 
In summary, our main contributions are as follows:
\begin{itemize}
    \item We introduce a new framework, SPEED, that uses Deferral Classifiers (DCs) to overcome the overconfidence issue in EEDNNs.

    \item We develop a new strategy to train DCs using training samples that are labeled based on their confidence scores across the layers of DNN.  

    \item Our method generalizes well to various domains with minimal loss in performance, making it robust to domain shifts during inference.

    \item We perform a theoretical analysis to provide a condition under which the model will be within a given risk limit.
    
\end{itemize}

\section{Related works}

\textbf{Early Exit (EE) techniques} allow models to make early predictions based on input complexity, have been widely applied across various domains, including image classification with various ways to choose a layer for exiting \cite{teerapittayanon2016branchynet, huang2017multi, kaya2019shallow, wang2019dynexit, wolczyk2021zero}, natural language processing (NLP) \cite{xin2020deebert, bajpai-hanawal-2024-ceebert, zhou2020bert, bajpaibeem,  zhu2021leebert}, and image captioning \cite{bajpai-hanawal-2024-capeen, fei2022deecap, bajpai2025free, tang2023you}. These methods are known for their strong generalization capabilities, as demonstrated by models like CeeBERT \cite{bajpai-hanawal-2024-ceebert, bajpai-hanawal-2024-dadee}. A key differentiator among existing approaches lies in the choice of confidence measures, such as prediction consistency \cite{zhou2020bert}, ensemble methods \cite{sun2021early}, and output entropy \cite{liu2021elasticbert}. Additionally, different training strategies, including separate training \cite{xin2020deebert} and joint training \cite{zhou2020bert}, have been explored to enhance the EE methods. With better inference, they also have good distributed inference applications \cite{bajpai2024distributed, pacheco2024ucbee}.
EE techniques have also been applied in practical settings like distributed inference \cite{ bajpai2024splitee}, where models are deployed across devices with varying computational capacities, such as mobile, edge, and cloud environments. For a survey of EE methods in NLP see \cite{bajpai2025survey}.

\textbf{Selective Prediction (SP)}, also referred to as selective classification, has been extensively explored across domains. \cite{chow1970optimum} introduced a cost-based rejection model, analyzing the trade-off between errors and rejections. This concept has also been studied in the context of Support Vector Machines (SVMs) \cite{brinkrolf2018interpretable, hendrickx2024machine}, nearest neighbors \cite{hellman1970nearest}, and boosting \cite{cortes2016boosting}, demonstrating its versatility across various classification paradigms.
In the domain of NNs, \cite{lecun1989handwritten} introduced a rejection strategy based on output logits, comparing the highest and second-highest activated logits to guide rejection decisions. SelectiveNet \cite{geifman2019selectivenet} proposed a selective classification technique to achieve a target risk with a specified confidence-rate function, laying the groundwork for risk-controlled predictions.

Recent advancements in SP such as Deep Gamblers \cite{liu2019deep} incorporate an additional class for abstention. \cite{feng2022towards} critically examined the selection mechanisms of these models and highlighted their limitations.
Their findings suggest that the improved performance of these models is largely due to the optimization process leading to a more generalizable model, which in turn enhances SP performance.

Furthermore, several works have proposed deferral-based approaches, where the decision to predict or defer is determined by a cost function. These methods either assign prediction costs equal to the model loss or defer predictions at a user-defined cost \cite{okati2021differentiable, mozannar2023should, verma2023learning}.

Our approach differs from existing methods as: (1) Use of SP to improve the trustworthiness of EEDNNs has not been studied before to the best of our knowledge. (2) We use DCs to check the hardness of the samples, which aids in early exit decisions.  (3) We use separately trained DCs to preserve backbone optimality and improve robustness to domain shifts during inference.

\vspace{-0.15cm}

\section{Methodology}
We first highlight the issue of overconfidence in EEDNNs using the SST-2 dataset and discuss how Deferral Classifiers (DCs) can improve their inference time and trustworthiness. 

\begin{figure*}
\centering
  \includegraphics[scale=0.5]{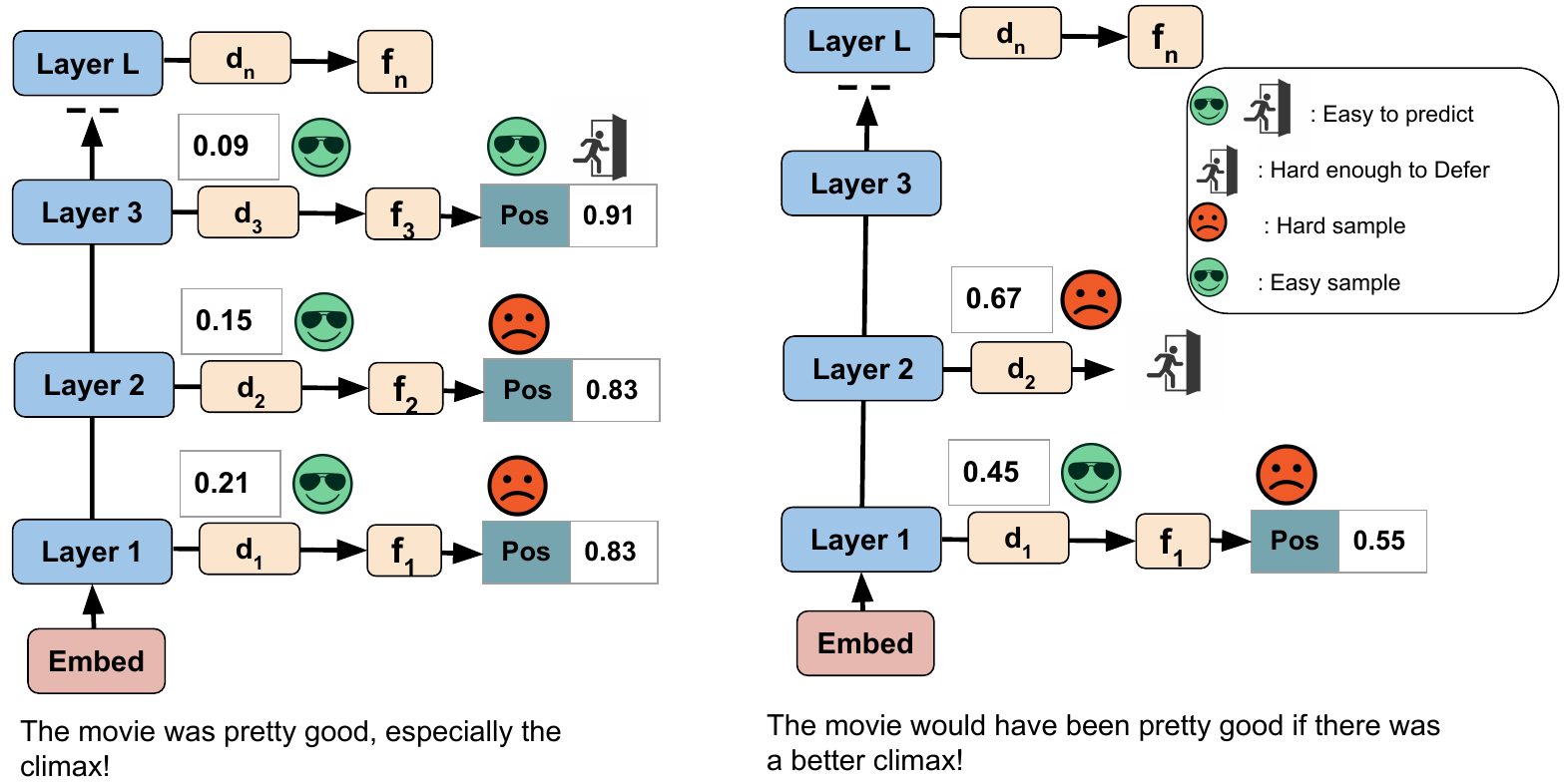}
 
\caption{The figure shows the effectiveness of our method during inference where an easy sample is exited early with a prediction a sample with fake confidence is detected and the model defers. The threshold for exiting is $0.90$ and the threshold for the DC score is $0.65$.}
 \label{fig:inference_example}
\end{figure*}

\subsection{Motivation}\label{sec: motivation}

To motivate our method, we show how the confidence values provided by the model can be misleading and reduce the trust in the model.  
In Fig~\ref{fig:average_confidence}, we show the average of confidence score and its variance for SST-2 samples recorded across all the exits on the true class of the trained Early Exit BERT. The samples are classified into three categories based on their confidence scores: 1) Confident/Easy: The model is confident over the true class. 2) Confused: The model sees similar scores in both classes and is unsure about the true class. 3) Fake: the model is overconfident (fake confidence) in the false class of these samples. 

Easy samples are the ones where the model is confident and the predictions are correct. Samples named confused are the ones where the model has ambiguity and is not sure about the true class, these samples can easily be detected and deferred to the expert. The main issue is with the samples that gain fake confidence. All the existing SP methods accept the prediction for these samples as they are convinced based on the confidence values and end up getting wrong predictions.


In our method, we explicitly train the DCs to catch such samples where the model has fake confidence. A high DC score indicates that it is a hard sample, while a high score of the exit classifier indicates it has high confidence in a class. Note that the samples with higher variance are the ones where the model is highly confused and fluctuates between the true and false classes. 

In Fig~\ref{fig:inference_example}, we contrast easy and fake samples and how the use of DCs helps in handling overconfidence. The sample on the left side of Fig \ref{fig:inference_example} is an easy sample. This is supported by low scores assigned by DCs and high scores by Exit Classifes (ECs). In the right one, the sample is hard. In this, the ECs assign high confidence (fake), but the DCs give high scores, indicating that the confidence is fake. Thus, with the use of DCs, the risk of accepting the fake confidence during the selective prediction is significantly reduced. We add a discussion for such samples in \ref{sec: fake_conf} and add more of these in Table \ref{tab: examples}.

The main challenge is training the DCs to identify the fake sample accurately. We use the Selective Prediction framework to build DCs.

\subsection{Selective Prediction Setting}
We begin with the general SP setting. Let $\mathcal{X}$ be the feature space of the dataset $\mathcal{D}$ and $\mathcal{Y}$ be the label set. Let $P(\mathcal{X}, \mathcal{Y})$ be the distribution over $\mathcal{X}\times \mathcal{Y}$. A model $f: \mathcal{X}\rightarrow \mathcal{Y}$ is called a prediction function and its true risk w.r.t. $P$ is $R(f) := E_{P(X, Y)}[l(f(x), y)]$ where $l: \mathcal{Y}\times\mathcal{Y}\rightarrow \mathbb{R}^{+}$ is any given loss such as the 0/1 loss. Given the labeled dataset $\mathcal{D} = \{(x_i, y_i)\}_{i=1}^{m} \subseteq (\mathcal{X}\times \mathcal{Y})$ sampled i.i.d. from $P(X, Y)$, the empirical risk of classifier $f$ is $\hat{r} = \frac{1}{m}\sum_{i = 1}^{m} l(f(x_i), y_i)$.

A selective model is a pair $(f, d)$ where $f$ is a prediction function and $d:\mathcal{X} \rightarrow \{0, 1\}$ is the deferral function that serves as the binary qualifier for $f$ as follows: 
\begin{equation}
\label{eq: general}
    (f, d)(x) := \left\{
        \begin{array}{ll}
            f(x) & \textit{if} \quad d(x) = 1\\
            \text{defer} & \textit{if} \quad d(x) = 0
        \end{array}
    \right.
\end{equation}
This equation shows the general setting of the selective models. The performance of a selective prediction model is quantified using coverage and risk. Coverage is defined as $\phi(g):=E_P[d(x)]$
 and the selective risk of $(f, d)$ is defined as: 
\begin{equation}
    R(f, d):=\frac{E_P[l(f(x), y)(d(x))]}{\phi(d)}
\end{equation}
Their empirical counterparts are given as $\hat{\phi}(d|\mathcal{D}):=\frac{1}{m}\sum_{i=1}^{m}d(x_i)$  and
\begin{equation}
\label{eqn: EmpRiskSP}
    \hat{r}(f, d|\mathcal{D}):=\frac{\sum_{i=1}^{m}l(f(x_i), y_i)(d(x_i))}{\hat{\phi}(d|\mathcal{D})},
\end{equation}
respectively. The goal is to find a pair $(f,d)$ such that  $ \hat{r}(f, d|\mathcal{D})$ is minimized. 


We adapt this setting to our case with EEDNNs.  We attached a Deferral Classifier (DC) in addition to the Early Exit classifier (EC) at each layer of DNN. Let $d_i$ and $f_i$ denote the DC and EC at the $i$th layers.  Their predictions are based on the confidence scores described as follows. 
For $x\sim \mathcal{D}$, let $\mathcal{P}_i(c)$ denote class probability that $x$ belongs to class $c\in \mathcal{C}$ by the EC at layer $i$. We define confidence score, denoted $C_i$, as the maximum over the class probability, i.e. $C_i:=\max_{c\in \mathcal{C}}\hat{\mathcal{P}}_i(c)$. 
The decision to exit with a prediction at the $i$th layer is made based on the value of $C_i$.  For a given threshold, $\alpha$ (exit threshold), if $C_i\geq\alpha$ the sample will be assigned a label $\hat{y} = \arg\max_{c\in\mathcal{C}}(\hat{\mathcal{P}}_i(c))$ and the sample is not further processed. $\alpha$ models the accuracy-efficiency trade-off.

For DCs, they output a score of hardness as the output of the linear layer, which takes hidden representations of the intermediary layer as input. Let $S_i$ denote the hardness score at the $i$th DC. For a given threshold $\beta$ (deferral threshold), a sample is deferred at the $i$th layer if $S_i\geq\beta$. $\beta$ models the risk-coverage trade-off. A higher value of $\beta$ will have a lower risk with lower coverage and vice versa. We define the function $d_i(x)$ as:
\begin{equation}
\label{eq: oue_specific}
    d_i(x) := \left\{
        \begin{array}{ll}
            1 & \textit{if} \quad  S_i<\beta\\
            0 &  \textit{otherwise}  
        \end{array}
    \right.
\end{equation}
Hence at every layer the tasks of the pair $(f_i, d_i)$ can be written as:
\begin{equation}
\label{eq: our_condition}
    (f_i, d_i)(x) := \left\{
        \begin{array}{ll}
            \text{defer} & \textit{if} \quad d_i(x) = 0, \\
            f_i(x) & \textit{if} \quad d_i(x) = 1 \\
                   & \quad \text{and } C_i \geq \alpha, \\
            i \gets i+1 & \textit{if} \quad d_i(x) = 1 \\
                       & \quad \text{and } C_i < \alpha.
        \end{array}
    \right.
\end{equation}

where $\alpha $ is the threshold that decides the early exiting of samples and $\beta$ named as the risk threshold is the hyperparameter that models the risk-coverage trade-off, the higher value of the risk factor increases the risk as well as coverage and vice versa.

When a sample is deferred it is exited from the DNN and assigned to an expert for prediction. The expert is the one who assigns the true label to the sample.
The coverage for the pair $(f_i, d_i)$ could be defined as $\phi(d_i):=E_P[d_i(x)|C_i\geq\alpha]$ and the total coverage across all the layers could be defined as  $\phi(G) = \sum_{i=1}^{n}\phi(d_i|C_i\geq\alpha)$ 
The selective risk of $(f_i, d_i)$ and the overall risk are defined as 
\begin{equation}
\label{eqn:RiskFinal}
    R_i(f_i, d_i):=\frac{E_P[l(f_i(x), y)(d_i(x))|C_i\geq\alpha]}{\phi(d_i)}. 
\end{equation}
\begin{equation}
    R = \sum_{i=1}^{n}R_i(f_i, d_i)
\end{equation}
Similar to \ref{eqn: EmpRiskSP}, we can define the empirical selective risk, and our goal is to minimize it over the possible values of $(\alpha, \beta)$. Note that the total risks discourage sending all samples to an expert and 
the pair $(\alpha, \beta)$
together decide the overall risk-accuracy trade-off of the model.

Our framework starts with a pre-trained DNN  and obtains an EEDNN with DCs using the following three steps. 1) Trains the ECs attached at each layer with frozen back-bone parameters. 2) Prepare the samples with appropriate labels to train the DCs. 3) Trains the DCs attached to each layer with frozen backbone and EC parameters. We next discuss each of these steps.

\subsection{Training Exit Classifiers}

Let $\mathcal{D}$ represent the data distribution with label space $\mathcal{C}$ used for training the backbone. The layers of the DNN are denoted as $L_1, L_2, \ldots, L_n$, with EC $f_i$ attached to layer $L_i$. For a given layer $L_i$, the hidden state $\mathbf{h}_i$ is computed as $\mathbf{h}_i = L_i(\mathbf{h}_{i-1})$, where $\mathbf{h}_0 = \textit{embedding}(x)$. Each EC maps the hidden representations to class probabilities, i.e., $y_i = f_i(\mathbf{h}_i)$. The loss for the $i$th exit classifier could be written as:
\begin{equation}
    \mathcal{L}_i = \mathcal{L}_{CE}(f_i(\mathbf{h}_i), y^*)
\end{equation}
where $y^*$ is the true label and $\mathcal{L}_{CE}$ denotes the cross-entropy loss. We learn the parameters for all the ECs simultaneously following the approach outlined by \cite{kaya2019shallow}, with the overall loss function defined as $\mathcal{L} = \frac{\sum_{i=1}^{n}i\cdot \mathcal{L}_i}{\sum_{i=1}^{n}i}$ where $n$ is the number of layers in the backbone. The weighted average considers the relative inference cost of each EC. Subsequently, the backbone parameters are trained along with all the EC  parameters.

\subsection{Training data for DCs}\label{sec:training_DCs}
We perform supervised training of DCs using samples that are labeled as easy and hard. To obtain the labels, we leverage the confidence score of ECs attached to the backbone. 

For a sample $x$ with label $y^{*}$, let $\mathcal{P}_i(y^{*}|x)$ denote the class probability $f_i$ assigns to $x$ on class $y^{*}$. The average true class probability score of the sample across all the layers is
$\hat{\mu}_x = \frac{1}{n}\sum_{i=1}^n \mathcal{P}_{i}(y^{*}|x)$, and the variability is $\hat{\sigma}_x = \frac{\sum_{i=1}^n(\mathcal{P}_i(y^*|x_i)-\hat{\mu_x})^2}{n}$. We sort the samples as per their scores $\hat{\mu}_x$ and label top-K\% of the sorted samples as hard and the remaining as easy ones. We do not use the variance as a metric since high variance is only for the samples that fluctuate and might not include the samples where the model is highly confident on the wrong class.
However, in the split, most of the high-variance samples are included as hard samples as samples with high variance often exhibit low overall confidence. 

The rationale behind creating these labels for DC training is that the hard samples might share some characteristics and patterns that can be learned and those patterns are usually shared across domains making our method generalizable.

\subsection{Training the DCs}
We attach the DCs at every layer of EEDNN after freezing the EEDNN parameters. The DCs are linear classifier layers whose task is to map the hidden representations at every layer to a score of hardness in the range of $0$ and $1$ where a higher score means a harder sample. The loss for the $i$th DC is:
\begin{equation}
    \mathcal{L}_i^{DC} = \mathcal{L}_{CE}(DC_i(h_i), z_i)
\end{equation}
where $z_i$ is the binary label of the sample with $0$ denoting easy and $1$ denoting hard sample. We learn the parameters of all the DCs simultaneously. The overall loss function could be written as $\mathcal{L}^{DC} = \frac{\sum_{i=1}^{n}i\cdot \mathcal{L}_i^{DC}}{\sum_{i=1}^{n}i}$. The higher weight to deeper layers is based on the intuition that the deeper layer's hidden representations have high-level knowledge of the sample providing more critical information to the DC at that layer.

Once the DC training is complete, its parameters are frozen. Using the validation data, the parameters $(\alpha, \beta)$ are chosen to minimize the overall empirical risk defined in (\ref{eqn:RiskFinal}). Note that in resource constraint scenarios, DCs can share the parameters across all the layers with a minor loss in performance (see Appendix \ref{sec: shared_params}). However,  having shared parameters only saves the storage space and not the inference cost, hence we use different DCs at different layers. As we use single linear layers as DCs, the computational cost is negligible. As compared to the full model size DC size is approximately 1e-5$\times$ of the full model adding up to negligible cost during inference. Even if we add up the parameters of all the DCs it will be only 8e-4$\times$ of the full model (see Appendix \ref{sec: shared_params}).

\subsection{Inference}
For a test input $x\sim \tilde{\mathcal{D}}$ where $\tilde{\mathcal{D}} = \mathcal{D}$ when the source and target domain are the same, and for a given thresholds $\alpha$ and $\beta$, let the sample is processed till the $i$th layer. At the $i$th layer, its $S_i$ value is calculated using DC. If $S_i\geq\beta$, the sample is deferred and sent to an expert that returns the gold label. If $S_i<\beta$, then the sample is referred to EC in the same layer, and its confidence score $C_i$ is checked. If $C_i\geq \alpha$, then the sample exits with a label completing the inference, else the sample is taken into the next layer, and the process continues till the sample reaches the last layer.

If the sample reaches the final layer, then the decision to infer or defer depends only on the confidence score of EC. If $C_i\geq\alpha$, then the sample is assigned a label. Otherwise, it defers the sample irrespective of the confidence of the DC. Hence, at every layer (except the final layer) of the backbone, every sample is either inferred, deferred, or passed on to the next layer. The following result establishes a connection between the error rate of DC and the risk of exit classifiers. 
\begin{lemma}\label{thm: lemma}
Consider an EEDNN with $n$ layers. Let $q_t$ denote the error rate of the $t$th exit classifier and $q_t^d$ be the error rate of the $t$th DC. Then the risk of the $t$th exit classifier will be less than $\gamma$ if the deferral classifier error rate is $q_t^d<\frac{1}{1+((1/\gamma)-1)(q_t/(1-q_t))}$. 
\end{lemma}
The condition established in the above Lemma is easy to satisfy. For instance, let us consider that $q_t =0.2$ and $\gamma = 0.1$ then the discriminator error rate should be less than $0.30$ i.e., the DC at $t$th layer have an accuracy of $0.70$ on average in classifying the incoming sample as easy or hard, which is not a very hard criterion if the deferral classifier has been trained properly with appropriate data. This further proves the importance of training the DCs and the critical importance of the training dataset used to train the DCs. Observe that the bound becomes tighter with an increase in value of $\gamma$ as well as when the value of $q_t$ increases. It shows that for a classifier making more mistakes, the discriminator should be better trained. 

\begin{theorem}\label{thm: theorem}
Consider an EEDNN with $n$ layers. Let $q_t$ denote the error rate of the $t$th exit classifier and $q_t^d$ be the error rate of the $t$th DC. Also let $q_{max} = \max \{q_1, q_2, \ldots, q_n\}$ and $q_{max}^d = \max\{q_1^d, q_2^d, \ldots, q_n^d\}$. Then the risk of the full EEDNN will be less than $\gamma$ if the maximum deferral classifier error rate is $q_{max}^d<\frac{1}{1+((1/\gamma)-1)(q_{max}/(1-q_{max}))}$.
\end{theorem}

This theorem gives a condition on the maximum error rate for a DC across all the layers such that the overall error rate of the EEDNN backbone is less than $\gamma$. Note that this condition could also be easily satisfied. For instance, consider $\gamma = 0.1$, and $q_{max} = 0.3$, then the bound is $q_{max}^d<0.21$, which means that all the DCs should have an error rate less than $0.21$. Again, if the DCs are efficiently trained, this criterion is not hard to meet. We can again make similar proportionality conditions on $\gamma$ and $q_{max}$ as done for Lemma \ref{thm: lemma}.

\section{Experiments}
In this section, we provide details of the experimental setup, key findings and analysis of our work.
\begin{table*}[]
\centering
\small
\begin{tabular}{lcc|cccc|c}
\hline
\textbf{Model}          & \textbf{BERT} & \textbf{PBEE}  & \textbf{BERT-SR} & \textbf{SelNet} & \textbf{Cal-RF}  & \textbf{PBEE-SR} & \textbf{Ours} \\ \hline
\textbf{Data}           & \multicolumn{6}{c}{\textbf{Risk/Coverage}}                                                               \\ \hline
SST2                    & 06.3/100  & 07.9/100       & 04.7/91.8        & 04.1/91.5       & 04.6/94.5       &   
 05.0/\textbf{95.6}         & \textbf{03.5}/95.4     \\
IMDB                    & 10.4/100   &  10.3/100    & 05.2/86.4        & 04.8/85.3       & 04.1/85.1       & 05.5/86.7         & \textbf{03.9}/\textbf{87.6}     \\
Yelp                    & 07.2/100  &   06.8/100  & 02.1/80.0        & 01.9/79.2       & 01.9/78.9       & 02.1/81.4         & \textbf{01.7}/\textbf{82.3}     \\
SciTail                 & 09.7/100   &    09.5/100    & 03.3/84.5        & 03.5/84.0       & 03.1/83.8       & 03.7/85.2         & \textbf{03.1}/\textbf{86.5}     \\
MRPC                    & 14.5/100     &    14.0/100   & 08.6/72.7        & 08.2/73.2       & 07.2/70.8       & 10.7/\textbf{75.3}         & \textbf{06.5}/74.9     \\
QQP                     &  10.7/100      &  10.4/100       &    06.4/87.4              &     05.8/86.8            &   06.1/87.9              &      06.6/88.5             &    \textbf{05.6}/\textbf{89.5}                \\
MNLI                    & 15.5/100   &  16.6/100     & 09.3/86.3        & 08.9/85.2       & 08.8/84.7       & 09.7/87.0         & \textbf{08.3}/\textbf{87.2}     \\
SNLI                    & 10.7/100     &  11.0/100  & 07.9/82.4        & 07.5/81.8       & 07.3/80.3       & 08.2/83.1         & \textbf{06.5}/\textbf{84.3}     \\ \hline
\textbf{Avg. Risk/Cov.} & 10.6/100      &  10.8/100   & 06.3/84.2         & 05.9/83.4        & 05.8/83.7       & 06.8/85.7         & \textbf{04.9}/\textbf{86.0}      \\
\textbf{Avg.Speed}      & 1.00x       &  1.71x  & 1.00x            & 1.00x           & 1.00x           & 1.71x             & \textbf{2.05x}         \\ \hline
\end{tabular}
\caption{In-Domain results: Results on the BERT backbone where the test and train set have the same distribution. We report the risk, coverage, average risk (Avg. Risk), average coverage (Avg. Cov.) and average speedup.}
\label{tab: ID-res}

\end{table*}

\vspace{-0.3cm}
\subsection{Datasets}
We utilized most of the GLUE \cite{wang2019glue} and the ELUE \cite{liu2021elasticbert} datasets. We evaluated \algo{} on different datasets covering three types of classification tasks. The datasets used for evaluation are:
\noindent
\textbf{1) Sentiment classification:} IMDB is a movie review classification dataset and Yelp consists of reviews from various domains such as hotels, restaurants etc. iii) SST-2 is also a similar type of dataset with the sentiment analysis task.
\noindent
\textbf{2) Entailment classification:} We have used the SciTail dataset created from multiple questions from science exams an web sentences. MRPC (Microsoft Research Paraphrase Corpus) dataset which also has a semantic equivalence classification task of a sentence pair extracted from online news sources. We also perform experiment on the QQP (Quora Question Pair) dataset. 
\noindent
\textbf{3) Natural Language Inference task:} We have used the MNLI and SNLI datasets for NLI tasks. SNLI is a collection of human-written English sentence pairs labeled for classification with labels \textit{entailment, contradiction} and \textit{neutral}.

In the Appendix \ref{tab:image_res}, we also include image datasets such as CIFAR-10 and Caltech-256 for image classification tasks.

\subsection{Experimental setup}

\textbf{i) Training the EE backbone on the source dataset:} Initially, we train the backbone on the train split of the source dataset. We add a linear output layer after each intermediate layer of the BERT/RoBERTa model whose task is to map the hidden representation to class probabilities. We run the model for $5$ epochs. We perform a grid search over batch size of $\{8, 16, 32\}$ and learning rates of \{1e-5, 2e-5, 3e-5, 4e-5, 5e-5\} with Adam \cite{kingma2014adam} optimizer. We apply an early stopping mechanism and select the model with the best performance on the development set. 

\textbf{ii) Creating the dataset for DC training:} After training the EE backbone, we freeze the parameters of the backbone and then calculate the average confidence and variance of confidence across all the exit points. The dataset is then sorted in ascending order of confidence. The top-K\% samples are provided the hard label and the remaining samples are treated as the easy ones. We choose K$=33$ for better balance between easy and hard samples so that the DCs are not biased towards one class, however, an ablation study has been made with different $K$ values in Appendix \ref{sec: top-k}. After creating the dataset, we attach the DCs i.e., a single linear layer mapping the hidden representations to hardness score. A hard sample will have a score closer to 1 and an easier sample close to 0. 

The training for DCs is performed for additional $3$ epochs. After training the DCs, the model can classify or defer early at each layer. For training the hyperparameters $\alpha$ we perform a grid search over the set \{0.75, 0.8, 0.85, 0.9, 0.95\} and for $\beta$, we choose the set as \{0.55, 0.6, 0.65, 0.7, 0.75\}. The thresholds could be chosen based on the user requirements of risk and coverage. We set the thresholds as the ones that show similar trade-offs for comparison with existing baselines. We plot the risk-coverage trade-off and discuss in the Appendix \ref{sec:risk_cov} (see Figure \ref{fig:iid_tradeoff}, \ref{fig:ood_tradeoff}). We also show the behaviour of the model over changing values of $\alpha$ and $\beta$ in Table \ref{tab:ablation_alpha_beta}. The experiments are conducted on NVIDIA RTX 2070 GPU with an average overall runtime of $\sim 3$ hours and a maximum run time of $\sim 10$ hours for the MNLI dataset. For more details see Appendix \ref{sec: computation_complexity} where we discuss the computational complexity and generalization of our method.

\textbf{iii) Inference:} We perform the inference on the test split of the source dataset and to show the generalization capabilities of our model, we also perform inference on a target domain dataset. Also, to maintain consistency with previous methods, we use the speed-up ratio as the metric to assess our model's improvement in speed as compared to existing methods. The speed-up ratio could be written as: $\frac{\sum_{i = 1}^n n\times x_i}{\sum_{i = 1}^n i\times x_i}$ where $x_i$ denotes the number of samples exiting from the $i$th layer and $n$ denotes the number of layers in  the backbone. Observe that $x_i$ consists of the samples that either get an early inference or early deferral. Also, note that the risk can be converted to accuracy on the covered samples and to the overall accuracy.

\begin{table*}[]
\centering
\small
\begin{tabular}{lcc|cccc|c}
\hline
\textbf{Data/Model}     & \textbf{BERT} &  
\textbf{PBEE}&  \textbf{BERT-SR} & \textbf{SelNet} & \textbf{Cal-RF} & \textbf{PBEE-SR} & \textbf{Ours}      \\ \hline
Src. - Tgt.                  & \multicolumn{6}{c}{Risk/Cov.}                                                                                 \\ \hline
SST-IMDB                & 19.8/100    &  20.2/100 & 14.9/86.0        & 14.2/85.1       & 14.6/87.0       & 14.8/86.4         & \textbf{12.8}/\textbf{87.2}          \\
SST-Yelp                & 17.9/100   &   18.7/100     & 11.0/78.3        & 11.4/79.0       & 11.9/80.4       & 13.7/81.7         & \textbf{09.0}/\textbf{82.0}          \\
IMDB-SST                & 18.1/100   &    18.5/100    & 15.5/85.2        & 14.8/86.0       & 15.1/85.5       & 16.2/86.2         & \textbf{13.8}/\textbf{85.9}          \\
IMDB-Yelp               & 24.3/100     &  25.4/100  & 15.2/81.5        & 15.0/79.8       & 14.6/80.8       & 15.5/80.1         & \textbf{13.5}/\textbf{82.1}          \\
Yelp-IMDB               & 22.8/100   &   23.2/100     & 13.2/73.4        & 12.6/72.7       & 12.9/72.7       & 14.8/73.9         & \textbf{09.5}/\textbf{74.5}          \\
Yelp-SST                & 25.1/100    &  25.8/100   & 13.8/71.5        & 14.0/72.1       & 13.6/70.6       & 14.2/73.8         & \textbf{12.0}/\textbf{75.3}          \\
SNLI-MNLI               & 25.1/100   &  26.5/100     & 14.9/80.7        & 14.5/79.2       & 13.9/78.5       & 15.2/81.6         & \textbf{12.3}/\textbf{82.8}          \\
MNLI-SNLI               & 19.8/100    &  20.6/100    & 11.4/76.8        & 10.8/74.8       & 11.3/76.2       & 12.6/\textbf{77.9}         & \textbf{09.5}/77.4          \\
MRPC-SciTail            & 35.7/100   &    36.8/100   & 25.6/74.2        & 24.1/71.6       & 25.9/\textbf{75.4}       & 21.5/72.8         & \textbf{18.8}/73.5          \\
SciTail-MRPC            & 29.9/100    &    30.7/100    & 23.3/69.9        & 22.7/67.8       & 23.1/69.2       & 24.7/70.3         & \textbf{21.4}/\textbf{70.9}          \\ \hline
\textbf{Avg. Risk/Cov.} & 21.3/100   & 24.6/100   & 15.8/77.7        & 15.4/76.8       & 15.7/77.6       & 16.3/78.4         & \textbf{13.2/79.1} \\
\textbf{Avg.Speed}      & 1.00x     &     1.57x     & 1.00x            & 1.00x           & 1.00x           & 1.57x             & \textbf{1.98x}     \\ \hline
\end{tabular}
\caption{Out-Of-Domain results: Results on the BERT backbone where the test set has a different distribution from the training set (Src. (source)- Tgt. (target)). We report risk, coverage, average risk and coverage (Avg. Risk/Cov.), and average speedup (Avg. Speed).}
\label{tab: OOD-res}
\end{table*}

\subsection{Baselines}
In this section, we detail the various baselines:

\textbf{1) BERT:}\cite{devlin2018bert} This baseline is where we perform vanilla BERT inference, where there is no selective prediction or early exiting. Hence the coverage in this case will always be full.

\textbf{2) PBEE:}\cite{zhou2020bert} It is the PABEE method applied to BERT, in this case also there is no selective prediction hence coverage will be full.

\textbf{2) BERT-SR:}\cite{geifman2017selective} It uses softmax-response where at the final output of the BERT model, the softmax layer is added after the final layer of the BERT and if the model is confident enough on the prediction, then it infers the sample, else it abstains. 

\textbf{3) SelectiveNet:}\cite{geifman2019selectivenet} In this there is an additional loss component to lower the risk of the model, however the criteria to abstain is similar to the BERT-SR method.

\textbf{4) Calibrator model:}\cite{pugnana2023model} This baseline considers a calibrator i.e., the output of the model is passed to a different smaller model, in our case, we use a random forest to decide about abstaining the sample. 

\textbf{5) PABEE SR:} We attach the softmax-response to the PABEE model, where the exits are attached, and there is a softmax layer at the final layer used to decide to abstain or perform inference.

For a fair comparison, the risk-coverage trade-off hyperparameter was chosen based on the values that lower the risk over validation data for baselines. The impact of not using DCs is similar to PABEE-SR baseline.

\begin{table*}[]
\centering
\small
\begin{tabular}{lccc}
\hline
\multicolumn{4}{c}{\textbf{IMDB}}                                                                                                                                  \\ \hline
\textbf{Example}                                                               & \textbf{True lbl.} & \textbf{Fake Avg. Conf.} & \textbf{DC} \\
I was really excited about this film, but I was wrong.                         & negative            & 0.96                                    & 0.87              \\
His last film was better than this.            & negative            & 0.89                                    & 0.82              \\
Low budget, but creepy enough to hold your interest                      & positive            & 0.91                                    & 0.84              \\
Don't judge it by bad cover picture. & positive            & 0.88                                    & 0.89              \\ \hline
\multicolumn{4}{c}{\textbf{SST-2}}                                                                                                                                 \\ \hline
mysterious and brutal nature                                                   & positive            & 0.90                                     & 0.91              \\
the under-7 crowd                                                              & negative            & 0.89                                    & 0.79              \\
seems endless                                                                  & negative            & 0.85                                    & 0.86              \\
a movie with two stars                                                         & positive            & 0.83                                    & 0.85              \\ \hline
\end{tabular}
\caption{Examples of samples achieving fake confidence, the table shows an example, its true label (True lbl.), the confidence of the model on the wrong class (Fake Avg. Conf.) and the deferral classifier score that abstains the sample early.}
\label{tab: examples}
\end{table*}

\begin{figure*}[htbp]
    \centering
    \subfloat[SST2-SST2 (i.i.d)]{
        \includegraphics[width=0.41\linewidth]{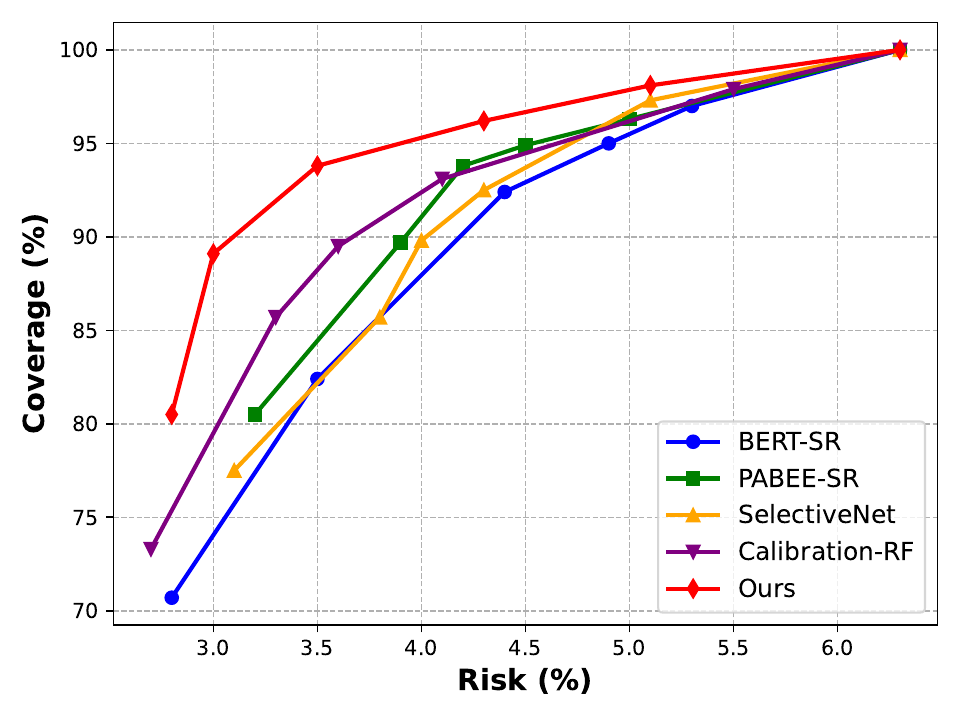}
        \label{fig:iid_tradeoff}
    }
    \hspace{0.02\linewidth}
    \subfloat[SST2-IMDB (o.o.d)]{
        \includegraphics[width=0.41\linewidth]{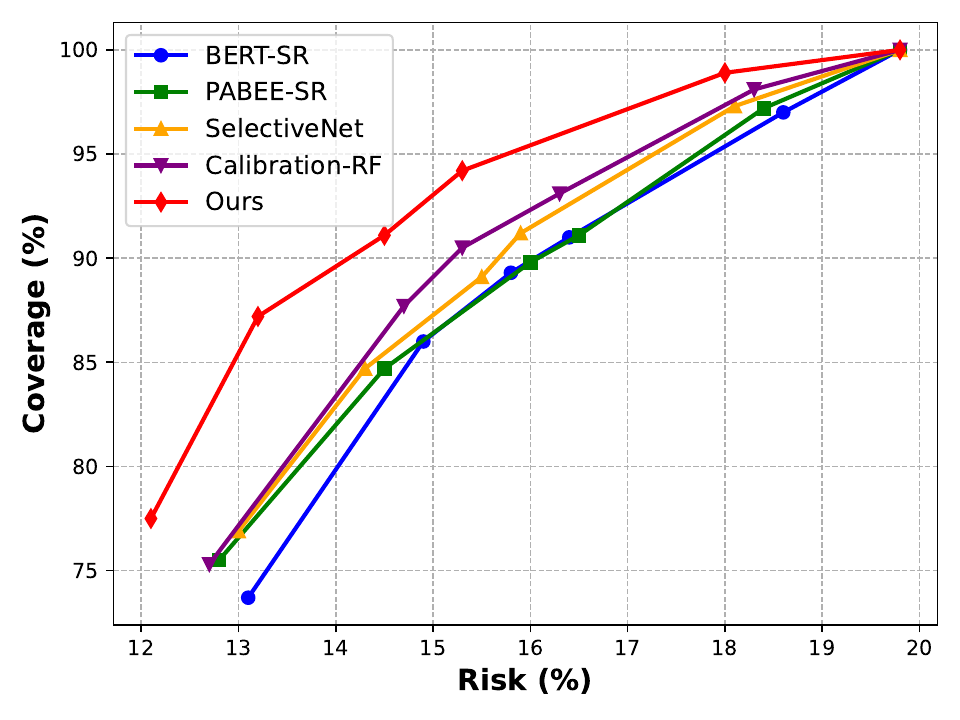}
        \label{fig:ood_tradeoff}
    }
    \caption{Risk-coverage trade-offs for in-domain and out-of-domain settings.}
    \label{fig:tradeoff_combined}
\end{figure*}

\subsection{Experimental results}
In Tables \ref{tab: ID-res} and \ref{tab:res_roberta} (Table 4 in Appendix), we present the average results of five runs on the test split of the training set using BERT and RoBERTa backbones, with stability reported in Appendix \ref{tab:staility_res}. Our method outperforms existing baselines, achieving lower risk by identifying and abstaining from samples with fake confidence early, unlike other methods that misclassify such samples (examples in Table \ref{tab: examples}, detailed discussion in Appendix \ref{sec: fake_conf}). Significant speedup is achieved by early exiting for easy samples and early deferral of hard samples, resulting in a $2.05\times$ improvement over vanilla BERT, reducing computational cost.

In Table \ref{tab: OOD-res}, we demonstrate the robustness of our method across domains, with minimal performance drop and risk increase when tested on OOD datasets. This is attributed to DCs trained to handle easy and hard samples, enabling better generalization. The existing method has poor generalization as they consider the confidence to decide deferring. The confidence of the model is affected during domain change which results in poor generalization. During domain change, our method significantly outperforms existing baselines.

Models like BERT-SR, PABEE, PABEE-SR, and Cal-RF rely on final-layer confidence, which can lead to higher risk due to fake confidence. SelectiveNet’s lower performance stems from backbone parameter optimization affecting backbone optimality. In contrast, our method reduces overthinking by early inferring easy samples, improving coverage and performance, as overthinking in existing methods often leads to abstention on confident samples processed too deeply \cite{zhou2020bert}.

\subsection{Some examples of Fake Confidence}\label{sec: fake_conf}
In Table~\ref{tab: examples}, we highlight cases where the model is highly confident yet incorrect. For instance, “a movie with two stars” yields fluctuating confidences across 12 layers due to ambiguity between cast and rating, often leading to misclassification. Similarly, “Don't judge it by bad cover picture” triggers overconfidence in the negative class, with consistently low true-class confidence. Our Discriminator Classifiers (DCs), using hardness scores before confidence, effectively flag such hard samples.

\subsection{Risk-coverage trade-off}\label{sec:risk_cov}
In Figure \ref{fig:iid_tradeoff} and \ref{fig:ood_tradeoff}, we show the risk-coverage trade-off, where the values are obtained by changing the risk hyperparameter in our case and for the existing baselines, we use the hyperparameters used in their codebases to obtain the results. Figure \ref{fig:iid_tradeoff} shows the in-domain trade-off where the model is trained in the SST2 dataset and tested on the test split of the same dataset while in Figure \ref{fig:ood_tradeoff} we plot the trade-off when the model is trained on the SST2 dataset and tested on the IMDB. We can observe that our method has a significantly smaller drop in coverage, and the risk decreases. The drop in coverage is lower when the domain of the dataset changes for our method. Note that better performance of our method comes as multiple DCs help to effectively catch the samples where the model can gain fake confidence or is unable to gain confidence. Also, mitigation of overthinking through EEs helps improve coverage pushing multiple metrics simultaneously.

\section{Conclusion}
In this work, we introduced \algo{}, a selective prediction tailored to improve inference latency and overconfidence issues in EEDNNs. We introduce new classifiers named deferral classifiers (DCs) to identify if a sample is hard for a given layer and if the model can get overconfident about its false class. Thus, it helps the model to understand what it does not know, which is crucial to prevent the model from making a false prediction with high confidence! We develop a method to train the DCs to effectively identify if a sample is hard or easy to infer at a given layer. This also improves the generalization across domains making it robust to domain changes.

\bibliographystyle{IEEEtran}
\bibliography{ref}

\begin{thebibliography}{10}
\providecommand{\url}[1]{#1}
\csname url@samestyle\endcsname
\providecommand{\newblock}{\relax}
\providecommand{\bibinfo}[2]{#2}
\providecommand{\BIBentrySTDinterwordspacing}{\spaceskip=0pt\relax}
\providecommand{\BIBentryALTinterwordstretchfactor}{4}
\providecommand{\BIBentryALTinterwordspacing}{\spaceskip=\fontdimen2\font plus
\BIBentryALTinterwordstretchfactor\fontdimen3\font minus \fontdimen4\font\relax}
\providecommand{\BIBforeignlanguage}[2]{{%
\expandafter\ifx\csname l@#1\endcsname\relax
\typeout{** WARNING: IEEEtran.bst: No hyphenation pattern has been}%
\typeout{** loaded for the language `#1'. Using the pattern for}%
\typeout{** the default language instead.}%
\else
\language=\csname l@#1\endcsname
\fi
#2}}
\providecommand{\BIBdecl}{\relax}
\BIBdecl

\bibitem{kaur2022trustworthy}
D.~Kaur, S.~Uslu, K.~J. Rittichier, and A.~Durresi, ``Trustworthy artificial intelligence: a review,'' \emph{ACM computing surveys (CSUR)}, vol.~55, no.~2, pp. 1--38, 2022.

\bibitem{gawlikowski2023survey}
J.~Gawlikowski, C.~R.~N. Tassi, M.~Ali, J.~Lee, M.~Humt, J.~Feng, A.~Kruspe, R.~Triebel, P.~Jung, R.~Roscher \emph{et~al.}, ``A survey of uncertainty in deep neural networks,'' \emph{Artificial Intelligence Review}, vol.~56, no. Suppl 1, pp. 1513--1589, 2023.

\bibitem{kaya2019shallow}
Y.~Kaya, S.~Hong, and T.~Dumitras, ``Shallow-deep networks: Understanding and mitigating network overthinking,'' in \emph{International conference on machine learning}.\hskip 1em plus 0.5em minus 0.4em\relax PMLR, 2019, pp. 3301--3310.

\bibitem{zhou2020bert}
W.~Zhou, C.~Xu, T.~Ge, J.~McAuley, K.~Xu, and F.~Wei, ``Bert loses patience: Fast and robust inference with early exit,'' \emph{Advances in Neural Information Processing Systems}, vol.~33, pp. 18\,330--18\,341, 2020.

\bibitem{10.1145/3639856.3639873}
\BIBentryALTinterwordspacing
D.~J. Bajpai, V.~K. Trivedi, S.~L. Yadav, and M.~K. Hanawal, ``Splitee: Early exit in deep neural networks with split computing,'' in \emph{Proceedings of the Third International Conference on AI-ML Systems}, ser. AIMLSystems '23.\hskip 1em plus 0.5em minus 0.4em\relax New York, NY, USA: Association for Computing Machinery, 2024. [Online]. Available: \url{https://doi.org/10.1145/3639856.3639873}
\BIBentrySTDinterwordspacing

\bibitem{10622954}
D.~J. Bajpai, A.~Jaiswal, and M.~K. Hanawal, ``I-splitee: Image classification in split computing dnns with early exits,'' in \emph{ICC 2024 - IEEE International Conference on Communications}, 2024, pp. 2658--2663.

\bibitem{pacheco2021early}
R.~G. Pacheco, F.~D. Oliveira, and R.~S. Couto, ``Early-exit deep neural networks for distorted images: Providing an efficient edge offloading,'' in \emph{2021 IEEE Global Communications Conference (GLOBECOM)}.\hskip 1em plus 0.5em minus 0.4em\relax IEEE, 2021, pp. 1--6.

\bibitem{bajpai-hanawal-2024-ceebert}
\BIBentryALTinterwordspacing
D.~J. Bajpai and M.~Hanawal, ``{C}ee{BERT}: Cross-domain inference in early exit {BERT},'' in \emph{Findings of the Association for Computational Linguistics: ACL 2024}, L.-W. Ku, A.~Martins, and V.~Srikumar, Eds.\hskip 1em plus 0.5em minus 0.4em\relax Bangkok, Thailand: Association for Computational Linguistics, Aug. 2024, pp. 1736--1748. [Online]. Available: \url{https://aclanthology.org/2024.findings-acl.101/}
\BIBentrySTDinterwordspacing

\bibitem{chow1970optimum}
C.~Chow, ``On optimum recognition error and reject tradeoff,'' \emph{IEEE Transactions on information theory}, vol.~16, no.~1, pp. 41--46, 1970.

\bibitem{pugnana2024deep}
A.~Pugnana, L.~Perini, J.~Davis, and S.~Ruggieri, ``Deep neural network benchmarks for selective classification,'' \emph{arXiv preprint arXiv:2401.12708}, 2024.

\bibitem{teerapittayanon2016branchynet}
S.~Teerapittayanon, B.~McDanel, and H.-T. Kung, ``Branchynet: Fast inference via early exiting from deep neural networks,'' in \emph{2016 23rd International Conference on Pattern Recognition (ICPR)}.\hskip 1em plus 0.5em minus 0.4em\relax IEEE, 2016, pp. 2464--2469.

\bibitem{huang2017multi}
G.~Huang, D.~Chen, T.~Li, F.~Wu, L.~Van Der~Maaten, and K.~Q. Weinberger, ``Multi-scale dense networks for resource efficient image classification,'' \emph{arXiv preprint arXiv:1703.09844}, 2017.

\bibitem{wang2019dynexit}
M.~Wang, J.~Mo, J.~Lin, Z.~Wang, and L.~Du, ``Dynexit: A dynamic early-exit strategy for deep residual networks,'' in \emph{2019 IEEE International Workshop on Signal Processing Systems (SiPS)}.\hskip 1em plus 0.5em minus 0.4em\relax IEEE, 2019, pp. 178--183.

\bibitem{wolczyk2021zero}
M.~Wo{\l}czyk, B.~W{\'o}jcik, K.~Ba{\l}azy, I.~T. Podolak, J.~Tabor, M.~{\'S}mieja, and T.~Trzcinski, ``Zero time waste: Recycling predictions in early exit neural networks,'' \emph{Advances in Neural Information Processing Systems}, vol.~34, pp. 2516--2528, 2021.

\bibitem{xin2020deebert}
J.~Xin, R.~Tang, J.~Lee, Y.~Yu, and J.~Lin, ``Deebert: Dynamic early exiting for accelerating bert inference,'' \emph{arXiv preprint arXiv:2004.12993}, 2020.

\bibitem{bajpaibeem}
D.~J. Bajpai and M.~K. Hanawal, ``Beem: Boosting performance of early exit dnns using multi-exit classifiers as experts,'' in \emph{The Thirteenth International Conference on Learning Representations}.

\bibitem{zhu2021leebert}
W.~Zhu, ``Leebert: Learned early exit for bert with cross-level optimization,'' in \emph{Proceedings of the 59th Annual Meeting of the Association for Computational Linguistics and the 11th International Joint Conference on Natural Language Processing (Volume 1: Long Papers)}, 2021, pp. 2968--2980.

\bibitem{bajpai-hanawal-2024-capeen}
\BIBentryALTinterwordspacing
D.~J. Bajpai and M.~K. Hanawal, ``{C}ap{EEN}: Image captioning with early exits and knowledge distillation,'' in \emph{Findings of the Association for Computational Linguistics: EMNLP 2024}, Y.~Al-Onaizan, M.~Bansal, and Y.-N. Chen, Eds.\hskip 1em plus 0.5em minus 0.4em\relax Miami, Florida, USA: Association for Computational Linguistics, Nov. 2024, pp. 6458--6472. [Online]. Available: \url{https://aclanthology.org/2024.findings-emnlp.376/}
\BIBentrySTDinterwordspacing

\bibitem{fei2022deecap}
Z.~Fei, X.~Yan, S.~Wang, and Q.~Tian, ``Deecap: Dynamic early exiting for efficient image captioning,'' in \emph{Proceedings of the IEEE/CVF Conference on Computer Vision and Pattern Recognition}, 2022, pp. 12\,216--12\,226.

\bibitem{bajpai2025free}
D.~J. Bajpai and M.~K. Hanawal, ``Free: Fast and robust vision language models with early exits,'' \emph{arXiv preprint arXiv:2506.06884}, 2025.

\bibitem{tang2023you}
S.~Tang, Y.~Wang, Z.~Kong, T.~Zhang, Y.~Li, C.~Ding, Y.~Wang, Y.~Liang, and D.~Xu, ``You need multiple exiting: Dynamic early exiting for accelerating unified vision language model,'' in \emph{Proceedings of the IEEE/CVF Conference on Computer Vision and Pattern Recognition}, 2023, pp. 10\,781--10\,791.

\bibitem{bajpai-hanawal-2024-dadee}
\BIBentryALTinterwordspacing
D.~J. Bajpai and M.~K. Hanawal, ``{DA}d{EE}: Unsupervised domain adaptation in early exit {PLM}s,'' in \emph{Findings of the Association for Computational Linguistics: EMNLP 2024}, Y.~Al-Onaizan, M.~Bansal, and Y.-N. Chen, Eds.\hskip 1em plus 0.5em minus 0.4em\relax Miami, Florida, USA: Association for Computational Linguistics, Nov. 2024, pp. 6389--6400. [Online]. Available: \url{https://aclanthology.org/2024.findings-emnlp.371/}
\BIBentrySTDinterwordspacing

\bibitem{sun2021early}
T.~Sun, Y.~Zhou, X.~Liu, X.~Zhang, H.~Jiang, Z.~Cao, X.~Huang, and X.~Qiu, ``Early exiting with ensemble internal classifiers,'' \emph{arXiv preprint arXiv:2105.13792}, 2021.

\bibitem{liu2021elasticbert}
\BIBentryALTinterwordspacing
X.~Liu, T.~Sun, J.~He, L.~Wu, X.~Zhang, H.~Jiang, Z.~Cao, X.~Huang, and X.~Qiu, ``Towards efficient {NLP:} {A} standard evaluation and {A} strong baseline,'' 2021. [Online]. Available: \url{https://arxiv.org/abs/2110.07038}
\BIBentrySTDinterwordspacing

\bibitem{bajpai2024distributed}
D.~J. Bajpai and M.~K. Hanawal, ``Distributed inference on mobile edge and cloud: An early exit based clustering approach,'' \emph{arXiv preprint arXiv:2410.05338}, 2024.

\bibitem{pacheco2024ucbee}
R.~G. Pacheco, D.~J. Bajpai, M.~Shifrin, R.~S. Couto, D.~S. Menasch{\'e}, M.~K. Hanawal, and M.~E.~M. Campista, ``Ucbee: A multi armed bandit approach for early-exit in neural networks,'' \emph{IEEE Transactions on Network and Service Management}, 2024.

\bibitem{bajpai2024splitee}
D.~J. Bajpai, A.~Jaiswal, and M.~K. Hanawal, ``I-splitee: Image classification in split computing dnns with early exits,'' \emph{arXiv preprint arXiv:2401.10541}, 2024.

\bibitem{bajpai2025survey}
D.~J. Bajpai and M.~K. Hanawal, ``A survey of early exit deep neural networks in nlp,'' \emph{arXiv preprint arXiv:2501.07670}, 2025.

\bibitem{brinkrolf2018interpretable}
J.~Brinkrolf and B.~Hammer, ``Interpretable machine learning with reject option,'' \emph{at-Automatisierungstechnik}, vol.~66, no.~4, pp. 283--290, 2018.

\bibitem{hendrickx2024machine}
K.~Hendrickx, L.~Perini, D.~Van~der Plas, W.~Meert, and J.~Davis, ``Machine learning with a reject option: A survey,'' \emph{Machine Learning}, vol. 113, no.~5, pp. 3073--3110, 2024.

\bibitem{hellman1970nearest}
M.~E. Hellman, ``The nearest neighbor classification rule with a reject option,'' \emph{IEEE Transactions on Systems Science and Cybernetics}, vol.~6, no.~3, pp. 179--185, 1970.

\bibitem{cortes2016boosting}
C.~Cortes, G.~DeSalvo, and M.~Mohri, ``Boosting with abstention,'' \emph{Advances in Neural Information Processing Systems}, vol.~29, 2016.

\bibitem{lecun1989handwritten}
Y.~LeCun, B.~Boser, J.~Denker, D.~Henderson, R.~Howard, W.~Hubbard, and L.~Jackel, ``Handwritten digit recognition with a back-propagation network,'' \emph{Advances in neural information processing systems}, vol.~2, 1989.

\bibitem{geifman2019selectivenet}
Y.~Geifman and R.~El-Yaniv, ``Selectivenet: A deep neural network with an integrated reject option,'' in \emph{International conference on machine learning}.\hskip 1em plus 0.5em minus 0.4em\relax PMLR, 2019, pp. 2151--2159.

\bibitem{liu2019deep}
Z.~Liu, Z.~Wang, P.~P. Liang, R.~R. Salakhutdinov, L.-P. Morency, and M.~Ueda, ``Deep gamblers: Learning to abstain with portfolio theory,'' \emph{Advances in Neural Information Processing Systems}, vol.~32, 2019.

\bibitem{feng2022towards}
L.~Feng, M.~O. Ahmed, H.~Hajimirsadeghi, and A.~Abdi, ``Towards better selective classification,'' \emph{arXiv preprint arXiv:2206.09034}, 2022.

\bibitem{okati2021differentiable}
N.~Okati, A.~De, and M.~Rodriguez, ``Differentiable learning under triage,'' \emph{Advances in Neural Information Processing Systems}, vol.~34, pp. 9140--9151, 2021.

\bibitem{mozannar2023should}
H.~Mozannar, H.~Lang, D.~Wei, P.~Sattigeri, S.~Das, and D.~Sontag, ``Who should predict? exact algorithms for learning to defer to humans,'' in \emph{International conference on artificial intelligence and statistics}.\hskip 1em plus 0.5em minus 0.4em\relax PMLR, 2023, pp. 10\,520--10\,545.

\bibitem{verma2023learning}
R.~Verma, D.~Barrej{\'o}n, and E.~Nalisnick, ``Learning to defer to multiple experts: Consistent surrogate losses, confidence calibration, and conformal ensembles,'' in \emph{International Conference on Artificial Intelligence and Statistics}.\hskip 1em plus 0.5em minus 0.4em\relax PMLR, 2023, pp. 11\,415--11\,434.

\bibitem{wang2019glue}
A.~Wang, A.~Singh, J.~Michael, F.~Hill, O.~Levy, and S.~R. Bowman, ``{GLUE}: A multi-task benchmark and analysis platform for natural language understanding,'' 2019, in the Proceedings of ICLR.

\bibitem{kingma2014adam}
D.~P. Kingma and J.~Ba, ``Adam: A method for stochastic optimization,'' \emph{arXiv preprint arXiv:1412.6980}, 2014.

\bibitem{devlin2018bert}
J.~Devlin, M.-W. Chang, K.~Lee, and K.~Toutanova, ``Bert: Pre-training of deep bidirectional transformers for language understanding,'' \emph{arXiv preprint arXiv:1810.04805}, 2018.

\bibitem{geifman2017selective}
Y.~Geifman and R.~El-Yaniv, ``Selective classification for deep neural networks,'' \emph{Advances in neural information processing systems}, vol.~30, 2017.

\bibitem{pugnana2023model}
A.~Pugnana and S.~Ruggieri, ``A model-agnostic heuristics for selective classification,'' in \emph{Proceedings of the AAAI Conference on Artificial Intelligence}, vol.~37, no.~8, 2023, pp. 9461--9469.

\end{thebibliography}

\section{Appendix}

\subsection{Proof for Lemma \ref{thm: lemma}}

\textbf{Proof:} For simplicity, we prove it for the binary case. We denote the probability that the sample early exits with a misclassification at $t$th classifier as $p_t^{misc}$ and the probability that the early exits with a classification at the $t$th classifier irrespective of correct or wrong classification as $p_t^{exit}$. We denote the error rate of the classifier at $t$th layer as $q_t$ and the deferral classifier at $t$th layer as $q_t^d$.

We also denote the random variable $X_t$ where $X_t = 1$ means a sample has exited the backbone with a classification irrespective of whether it is correct or wrong. $X_t = 0$ if the sample is rejected or passed to the next layer.

$P(X_t=1) = P(S_1<\beta, \ldots, S_t<\beta).P(C_1<\alpha, \ldots, C_i\geq \alpha)$

To upper bound the risk we have to get a condition such that $$\frac{p_t^{misc}}{p_t^{exit}}<\gamma$$

By the total probability law, we know that

$$p_t^{exit} = P(X_t = 1|mc)P(mc)+P(X_t=1|cc)P(cc)$$

where $mc$ is misclassified sample and $cc$ is the correctly classified samples. Further, we can write 

$$p_t^{misc} = P(X_t = 1, mc) = P(X_t = 1|mc)P(mc)$$

Now we can write 

$$\frac{p_t^{misc}}{p_t^{exit}} = \frac{P(X_t = 1|mc)P(mc)}{P(X_t = 1|mc)P(mc)+P(X_t=1|cc)P(cc)}$$

$$\frac{P(X_t = 1|mc)P(mc)}{P(X_t = 1|mc)P(mc)+P(X_t=1|cc)P(cc)}<\gamma$$

$$\frac{P(X_t=1|cc)P(cc)}{P(X_t=1|mc)P(mc)}>\frac{1}{\gamma}-1$$

Now $P(X_t=1|mc)$ means that the sample has fake confidence due to which it has satisfied the condition $C_i\geq\alpha$ and the deferral classifier was not successful in catching it and predicted it as easy, its probability is $q_t^d$. Hence

\begin{multline}
 P(X_t=1|mc) = P(S_1<\beta, \ldots, S_t<\beta)\\.P(C_1<\alpha, \ldots, C_i\geq \alpha).q_t.q_t^d   
\end{multline}

Similarly, $P(X_t = 1|cc)$ means that the sample has true confidence and the deferral classifier has correctly predicted it.

\begin{multline}
    P(X_t=1|mc) = P(S_1<\beta, \ldots, S_t<\beta)\\.P(C_1<\alpha, \ldots, C_i\geq \alpha).(1-q_t).(1-q_t^d)
\end{multline}

Now we can write that

$$\frac{(1-q_t).(1-q_t^d)}{q_t.q_t^d}>\frac{1}{\gamma}-1$$

Solving for $q_t^d$, we get that:

$$q_t^d<\frac{1}{1+((1/\gamma)-1)(q_t/(1-q_t))}$$

Hence proved.

\subsection{Proof of Theorem \ref{thm: theorem}}
For simplicity, we prove it for the binary case. We denote the probability that the sample early exits with a misclassification at $t$th classifier as $p_t^{misc}$ and the probability that the early exits with a classification at the $t$th classifier irrespective of correct or wrong classification as $p_t^{exit}$. Also, let $p_{misc}$ be the overall misclassification rate of the model and $p_{exit}$ be the probability that the sample exits the system with a classification. We denote the error rate of the classifier at $t$th layer as $q_t$ and the deferral classifier at $t$th layer as $q_t^d$.

We also denote the random variable $X_t$ where $X_t = 1$ means a sample has exited the backbone with a classification irrespective of whether it is correct or wrong. $X_t = 0$ if the sample is rejected or passed to the next layer.

$P(X_t=1) = P(S_1<\beta, \ldots, S_t<\beta).P(C_1<\alpha, \ldots, C_i\geq \alpha)$

Now, the samples that have been misclassified could be written as:

$$p_{misc} = \sum_{t=1}^n p_t^{misc} $$ 
$$= \sum_{t=1}^n P(X_t=1,mc) = \sum_{t=1}^n P(X_t = 1|mc).P(mc)$$ 

\begin{multline}
    =\sum_{t=1}^n P(S_1<\beta,\ldots, S_t<\beta)(q_t^d)\\. P(C_1<\alpha,\ldots, C_t\geq\alpha)(q_t)
\end{multline}

Now the samples that have exited the backbone with a classification could be written as:
$$p_{exit} = \sum_{t=1}^n p_t^{exit} = \sum_{t=1}^n P(X_t=1) = $$ 
$$\sum_{t=1}^n P(X_t = 1|mc).P(mc)+P(X_t = 1|cc).P(cc)$$  

\begin{multline}
    =\sum_{t=1}^n P(S_1<\beta,\ldots, S_t<\beta)\\. P(C_1<\alpha,\ldots, C_t<\alpha).(q_t.q_t^d+(1-q_t).(1-q_t^d))
\end{multline}

We assume that $P(S_1<\beta, \ldots, S_t<\beta)$ and $P(C_1<\alpha, \ldots, C_t\geq\alpha)$ are constants denoted as $a$ and $b$ respectively. However, it might not be always true, but for simplicity, we assume it.

Now we can write that
$$\frac{p_{misc}}{p_{exit}} = \frac{a.b.\sum_{t=1}^{n}q_t.q_t^d}{a.b.\sum_{t=1}^n(1-q_t)(1-q_t^d)} $$ 

$$< \frac{n.q_{max}.q_{max}^d}{n.(1-q_{max})(1-q_{max}^d)}<\gamma$$

where $q_{max}=\max\{q_1, \ldots, q_n\}$ and $q_{max}^d=\max\{q_1^d, \ldots, q_n^d\}$. Solving for $q_{max}^d$ and following the steps from Lemma \ref{thm: lemma}, we obtain the bound.

$$q_{max}^d<\frac{1}{1+((1/\gamma)-1)(q_{max}/(1-q_{max}))}$$

\subsection{Results for the Image tasks}
In table \ref{tab:image_res}, we show the results of our model on the Cifar10 and Caltech-256 datasets on the MobileNet model. We attach exit classifiers in a similar fashion. In this setup, we first train the model with early exits and then the DCs are trained. After this step, the clean test set images are used for inference (shown as pristine). Then we add noise to the images of the test split and consider that as a domain change, where the noise added is the Gaussian blur. The level of Blur depends on the value of $\sigma$, higher the value of $\sigma$, the more blur the image is. This translates to the scenarios of domain change in autonomous driving where the environment factor might change the distribution of incoming samples.

The performance of our method is similar in the image domain as well where our method outperforms all the existing baselines with a significant margin. This further proves the effectiveness of our method.

\begin{table}[]
\centering
\begin{tabular}{ccccc}
\hline
K    & SST-2    & MNLI     & SciTail  & QQP      \\ \hline
10\% & 4.3/95.7 & 9.1/88.4 & 3.5/86.8 & 6.2/90.2 \\
33\% & 3.5/95.4 & 8.3/87.2 & 3.1/86.5 & 5.6/89.5 \\
50\% & 3.2/90.2 & 8.0/84.8 & 2.9/85.3 & 5.1/86.9 \\
66\% & 2.5/80.5 & 5.7/75.9 & 2.3/72.6 & 4.1/74.5 \\
90\% & 0.7/32.4 & 1.6/35.8 & 0.4/41.7 & 2.2/47.3 \\ \hline
\end{tabular}
\caption{The results by choosing top-K samples as hard based on average confidence}
\label{tab:top-k}
\end{table}

\begin{table}
\centering
\small
\begin{tabular}{lcccc}
\hline
\textbf{Data/Mdl} & \textbf{RT} & \textbf{RT-SR} & \textbf{PB-SR} & \textbf{Ours} \\ \hline
\multicolumn{5}{c}{\textbf{Risk/Coverage}} \\ \hline
SST2          & 7.8/100  & 6.2/92.7 & 6.8/93.8 & \textbf{4.5}/\textbf{94.1} \\
Yelp          & 5.9/100  & 5.1/93.2 & 5.4/\textbf{94.0} & \textbf{4.2}/93.9          \\
IMDB          & 10.1/100 & 7.5/91.5 & 7.1/92.1 & \textbf{6.8}/\textbf{92.6} \\
SciTail       & 8.5/100  & 7.0/\textbf{90.2} & 6.8/89.4 & \textbf{5.7}/\textbf{89.7}          \\ 
MRPC    &    13.3/100     &    9.6/87.5       &       10.1/88.6    &    \textbf{9.0}/\textbf{88.6} \\
QQP    &     9.8/100      &    8.4/91.9      &     8.1/93.2     &    \textbf{7.5}/\textbf{93.7}\\    
\hline
\textbf{Avg.}    &   9.2/100 
   &    7.3/91.6      &      7.3/91.8   &  \textbf{6.2}/\textbf{92.1}  \\
\textbf{Avg. Spd.}    &   1.00x 
   &    1.00x      &      1.73x   &  \textbf{2.19x} \\ \hline
\end{tabular}
\caption{Results over the RoBERTa backbone.}
\label{tab:res_roberta}
\end{table}

\begin{table}[]
\centering
\begin{tabular}{c|ccc}
\hline
\multicolumn{1}{l|}{\textbf{$\alpha$/$\beta$}} & \textbf{0.55} & \textbf{0.65} & \textbf{0.75} \\ \hline
\textbf{0.75}                            & 03.7/94.0     & 04.1/96.5     & 04.7/97.4     \\
\textbf{0.85}                            & 03.2/93.2     & 03.5/95.4     & 04.1/96.3     \\
\textbf{0.95}                            & 01.5/88.6     & 01.9/89.9     & 02.1/91.6     \\ \hline
\end{tabular}
\caption{Ablation over the $\alpha$ and $\beta$ values.}
\label{tab:ablation_alpha_beta}
\end{table}

\begin{table*}[]
\centering
\small
\begin{tabular}{lcc|cccc|c}
\hline
\textbf{Model}          & \textbf{BERT} & \textbf{PBEE}  & \textbf{BERT-SR} & \textbf{SelNet} & \textbf{Cal-RF}  & \textbf{PBEE-SR} & \textbf{Ours (Shared DC)} \\ \hline
\textbf{Data}           & \multicolumn{6}{c}{\textbf{Risk/Coverage}}                                                               \\ \hline
SST2                    & 06.3/100  & 07.9/100       & 04.7/91.8        & 04.1/91.5       & 04.6/94.5       &   
 05.0/\textbf{95.6}         & \textbf{04.0}/95.5     \\
IMDB                    & 10.4/100   &  10.3/100    & 05.2/86.4        & 04.8/85.3       & \textbf{04.1}/85.1       & 05.5/86.7         & 04.2/\textbf{87.4}     \\
Yelp                    & 07.2/100  &   06.8/100  & 02.1/80.0        & 01.9/79.2       & 01.9/78.9       & 02.1/81.4         & \textbf{02.1}/\textbf{82.0}     \\
SciTail                 & 09.7/100   &    09.5/100    & 03.3/84.5        & 03.5/84.0       & 03.1/83.8       & 03.7/85.2         & \textbf{03.3}/\textbf{86.7}     \\
MRPC                    & 14.5/100     &    14.0/100   & 08.6/72.7        & 08.2/73.2       & 07.2/70.8       & 10.7/\textbf{75.3}         & \textbf{06.8}/75.2     \\
QQP                     &  10.7/100      &  10.4/100       &    06.4/87.4              &     05.8/86.8            &   06.1/87.9              &      06.6/88.5             &    \textbf{05.7}/\textbf{89.4}                \\
MNLI                    & 15.5/100   &  16.6/100     & 09.3/86.3        & 08.9/85.2       & 08.8/84.7       & 09.7/87.0         & \textbf{08.6}/\textbf{87.5}     \\
SNLI                    & 10.7/100     &  11.0/100  & 07.9/82.4        & 07.5/81.8       & 07.3/80.3       & 08.2/83.1         & \textbf{07.1}/\textbf{83.9}     \\ \hline
\textbf{Avg. Risk/Cov.} & 10.6/100      &  10.8/100   & 06.3/84.2         & 05.9/83.4        & 05.8/83.7       & 06.8/85.7         & \textbf{05.2}/\textbf{85.9}      \\
\textbf{Avg.Speed}      & 1.00x       &  1.71x  & 1.00x            & 1.00x           & 1.00x           & 1.71x             & \textbf{1.96x}         \\ \hline
\end{tabular}
\caption{In-Domain results: Results on the BERT backbone where the test and train set have the same distribution. We report the risk, coverage, average risk (Avg. Risk), average coverage (Avg. Cov.) and average speedup. In this table, the parameters of the DCs are shared.}
\label{tab: shared_params}
\end{table*}

\subsection{Shared Parameters of the DCs}\label{sec: shared_params}
In this section, we discuss the effect when the attached DCs have shared parameters. In table \ref{tab: shared_params}, we show the results when the parameters of DCs are shared. Note that there is a drop in performance, but that drop is not as significant as the task of all the DCs i.e., to provide a score to the sample based on its hardness. However, the computation complexity during inference is the same as the case when the parameters are not shared. The issue of storage is solved with shared parameters as the model size is reduced.

The size of DCs with shared parameters is \textit{hidden size} while when the parameters are not shared, it is \textit{hidden size $\times$ number of layers}. For instance, for the BERT model, the number of additional parameters with shared DCs is only 768 while when the DCs are not shared then the number of parameters are 9216. Hence for models like BERT, the size of DCs is not that large even with shared or unshared parameters. As compared to the model size the size of DCs is negligibly small. It means they add negligible computational complexity.

\subsection{Impact of the Top-K\% choice of samples}\label{sec: top-k}
In table \ref{tab:top-k}, we show the results when different values of K are chosen i.e., the number of hard samples used to train the DCs are varied. Note that as the percentage of hard samples increases the DCs start predicting many samples as hard that reduces coverage significantly. We choose K$=33\%$ as it correctly balances the risk-coverage trade-off and also, it does not makes the DCs biased towards a single class.

\subsection{Computational complexity}\label{sec: computation_complexity}
In this section, we discuss the computational complexity added to the backbone due to the addition of the DCs. Note that the size of a single DC for the BERT backbone is hidden size times one as the output is only a score of hardness. Hence, even if the DCs do not have shared parameters, the size of all the DCs across all the layers will be hidden size times the number of layers i.e., 9216 which is negligibly small as compared to the size of the BERT model that has 110M parameters. Also, not all the DCs are utilized for every sample, as easy and hard samples, might exit earlier before utilizing all the DCs. Similar is the case with the ECs where the only change is that each EC has a hidden size times the number of classes. This makes our method applicable to large-size LMs as well as the DCs attached in our method will not scale up the computational complexity due to the simple structure of the DCs.


\begin{table*}[]
\centering
\begin{tabular}{lccccc}
\hline
\textbf{Data/Model}          & \textbf{MNet} & \textbf{MNet-SR} & \textbf{SelNet} & \textbf{PABEE-SR} & \textbf{Ours} \\ \hline
\multicolumn{6}{c}{Caltech-256 Risk/Coverage}                                                                         \\ \hline
Pristine                     & 21.7/100      & 16.5/89.8        & 15.3/87.7       & 15.9/88.5         & \textbf{14.8}/\textbf{89.3}     \\
$\sigma = 0.5$  & 24.8/100      & 17.2/\textbf{86.0}        & 16.8/85.4       & 16.3/85.1         & \textbf{15.9}/85.9     \\
$\sigma = 0.75$ & 26.2/100      & 19.8/84.6        & 18.5/82.6       & 19.0/84.9         & \textbf{17.1}/\textbf{85.1}     \\
$\sigma = 1.0$  & 28.4/100      & 21.2/82.3        & 20.9/\textbf{81.9}       & 20.4/81.3         & \textbf{18.5}/81.8     \\ \hline
\multicolumn{6}{c}{Cifar10 - Risk/Coverage}                                                                           \\ \hline
Pristine                     & 7.5/100       & 4.8/93.4         & 3.9/92.2        & 4.9/\textbf{95.8}          & \textbf{4.1}/95.3      \\
    $\sigma = 0.5$  & 10.1/100      & 8.3/91.9         & 6.2/89.6        & 7.6/92.6          & \textbf{5.9}/\textbf{93.5}      \\
$\sigma = 0.75$ & 14.9/100      & 10.7/89.0        & 9.4/88.1        & 9.8/88.9          & \textbf{8.5}/\textbf{89.4}      \\
$\sigma = 1.0$  & 21.7/100      & 16.2/83.9        & 14.5/82.4       & 15.2/84.8         & \textbf{13.8}/\textbf{85.2}     \\ \hline
\end{tabular}
\caption{Results on CIFAR10 and Caltech-256 datasets with different levels of noise on the MobileNet backbone. }
\label{tab:image_res}
\end{table*}

\begin{figure}
    \centering
    \includegraphics[scale=.38]{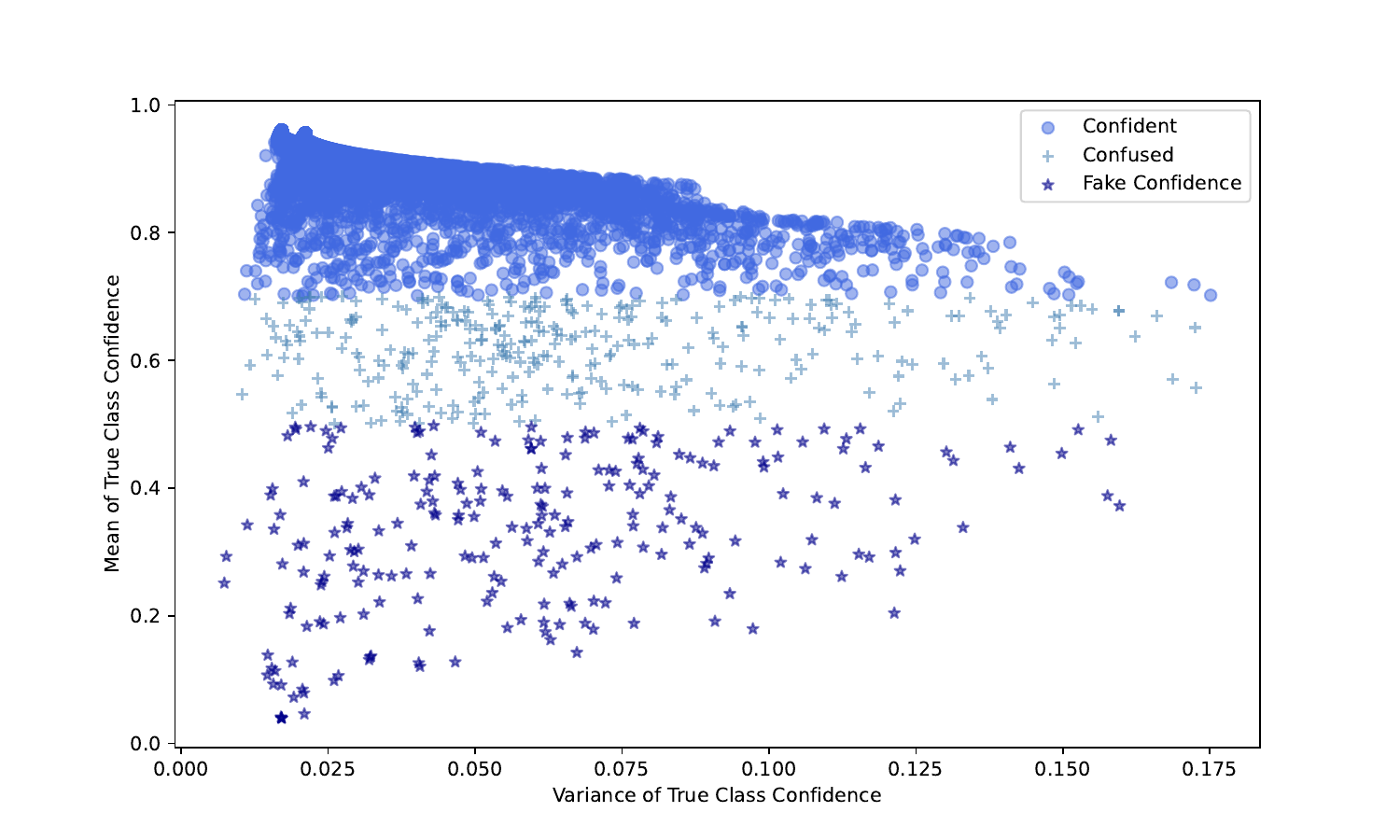}
    \caption{Average confidence values on IMDB dataset.}
  \label{fig:avg.conf_imdb}
\end{figure}

\begin{figure}
    \centering
    \includegraphics[scale=.5]{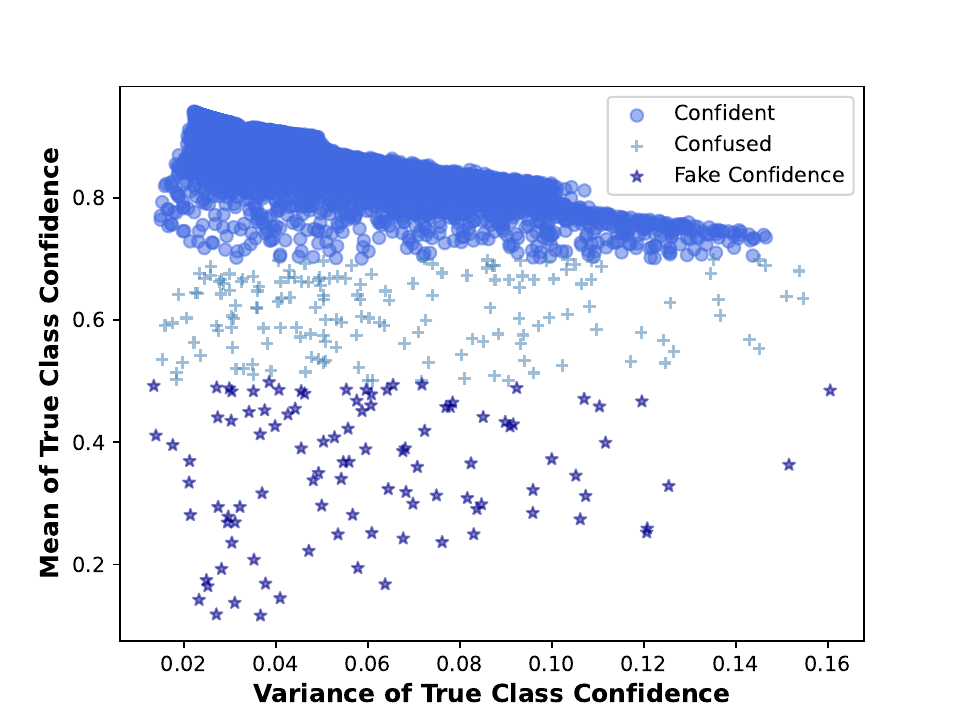}
    \caption{Average confidence values on SciTail dataset.}
  \label{fig:avg.conf_scitail}
\end{figure}


\begin{table*}[]
\centering
\begin{tabular}{lccccc}
\hline
\textbf{} & \textbf{BERT-SR}          & \textbf{SelNet}           & \textbf{Cal-RF}           & \textbf{PABEE-SR}          & \textbf{Ours}             \\ \hline
          & \multicolumn{5}{c}{Risk/Coverage}                                                                                                          \\ \hline
SST2      & 4.7$\pm$0.07 & 4.1$\pm$0.03 & 4.6$\pm$0.02 & 5.0$\pm$0.06  & 3.2$\pm$0.04 \\
IMDB      & 8.2$\pm$0.09 & 7.8$\pm$0.05 & 8.1$\pm$0.06 & 9.5$\pm$0.10  & 6.9$\pm$0.03 \\
Yelp      & 2.1$\pm$0.04 & 1.9$\pm$0.02 & 1.9$\pm$0.03 & 2.1$\pm$0.03  & 1.7$\pm$0.02 \\
SciTail   & 3.3$\pm$0.06 & 3.5$\pm$0.03 & 3.1$\pm$0.05 & 3.7$\pm$0.04  & 3.1$\pm$0.02 \\
MRPC      & 8.6$\pm$0.10 & 8.2$\pm$0.05 & 7.2$\pm$0.09 & 10.7$\pm$0.09 & 6.5$\pm$0.03 \\ \hline
\end{tabular}
\caption{Stability of our method as compared to others.}
\label{tab:staility_res}
\end{table*}

\end{document}